\documentclass[letterpaper, 10 pt, conference]{ieeeconf}  
\pdfoutput=1

 
 \makeatletter
 \newcommand*\titleheader[1]{\gdef\@titleheader{#1}}
 \AtBeginDocument{%
   \let\st@red@title\@title%
   \def\@title{%
     \bgroup\normalfont\large\centering\@titleheader\par\egroup
     \vskip0.5em\st@red@title}
 }
 \makeatother

\titleheader{\textit{Accepted Version}: IEEE/RSJ International Conference on Intelligent Robots and Systems (IROS), Oct.25-29, 2020 - Las Vegas, USA.
 \textsuperscript{\textcopyright} 2020 IEEE. Personal use of this material is permitted. Permission from IEEE must be obtained for all other uses. See \href{http://www.ieee.org/publications_standards/publications/rights/index.html}{\textit{here}} for more information.}
\usepackage[pscoord]{eso-pic}
\newcommand{\placetextbox}[3]{
\setbox0=\hbox{#3}
\AddToShipoutPictureFG{ \put(\LenToUnit{#1\paperwidth},\LenToUnit{#2\paperheight}){\vtop{{\null}\makebox[0pt][c]{#3}}}}
}
\placetextbox{.35}{0.05}{~\copyright IEEE All rights reserved. IEEE/RSJ IROS-2020, Oct.25-29. Las Vegas, USA.}

\IEEEoverridecommandlockouts                
\usepackage{epsfig} 
\usepackage{amsmath} 
\usepackage{amssymb}  
 \usepackage{subfigure}
\usepackage{algorithm}
\usepackage{algpseudocode}
\usepackage{amsmath}
\usepackage{etoolbox}
\makeatletter
\patchcmd{\@makecaption}
  {\scshape}
  {}
  {}
  {}
\makeatother
\usepackage[pdfusetitle]{hyperref}

\title{\LARGE \bf
Control Interface for Hands-free Navigation of Standing Mobility Vehicles based on Upper-Body Natural Movements
}
\author{Yang Chen$^{1}$, Diego {Paez-Granados}$^{2 \dagger}$, Hideki Kadone$^3$ and  Kenji Suzuki$^{4}$
 \thanks{$^\dagger$  is the corresponding author.}
\thanks{
$^{1}$ Y. Chen is with the School of Integrative and Global Majors (SIGMA), University of Tsukuba, Japan.
        {\tt\small chenyang@ai.iit.tsukuba.ac.jp}}  
\thanks{$^{2}$D. {Paez-Granados} is with the Learning Algorithms and Systems Laboratory (LASA), Ecole Polytechnique Federale de Lausanne (EPFL), Lausanne, Switzerland.
        {\tt\small dfpg@ieee.org}}  
\thanks{$^3$H. Kadone is with the Center for Innovative Medicine and Engineering, University of Tsukuba Hospital, Japan. 
       {\tt\small kadone@md.tsukuba.ac.jp}}  
\thanks{$^{4}$ K. Suzuki is with the Faculty of Engineering and Center for Cybernics Research, University of Tsukuba, Japan. 
        {\tt\small kenji@ieee.org}  
 	}
}
\begin{document}

\maketitle
\thispagestyle{empty}
\pagestyle{empty}


\begin{abstract} 			

In this paper, we propose and evaluate a novel human-machine interface (HMI) for controlling a standing mobility vehicle or person carrier robot, aiming for a hands-free control through upper-body natural postures derived from gaze tracking while walking. We target users with lower-body impairment with remaining upper-body motion capabilities. The developed HMI bases on a sensing array for capturing body postures; an intent recognition algorithm for continuous mapping of body motions to robot control space; and a personalizing system for multiple body sizes and shapes. We performed two user studies: first, an analysis of the required body muscles involved in navigating with the proposed control; and second, an assessment of the HMI compared with a standard joystick through quantitative and qualitative metrics in a narrow circuit task. We concluded that the main user control contribution comes from Rectus Abdominis and Erector Spinae muscle groups at different levels. Finally, the comparative study showed that a joystick still outperforms the proposed HMI in usability perceptions and controllability metrics, however, the smoothness of user control was similar in jerk and fluency. Moreover, users' perceptions showed that hands-free control made it more anthropomorphic, animated, and even safer.

\end{abstract}




\section{INTRODUCTION}\label{sec_intro}
\begin{figure}[t]
     	\centering
			\includegraphics[width=8cm]{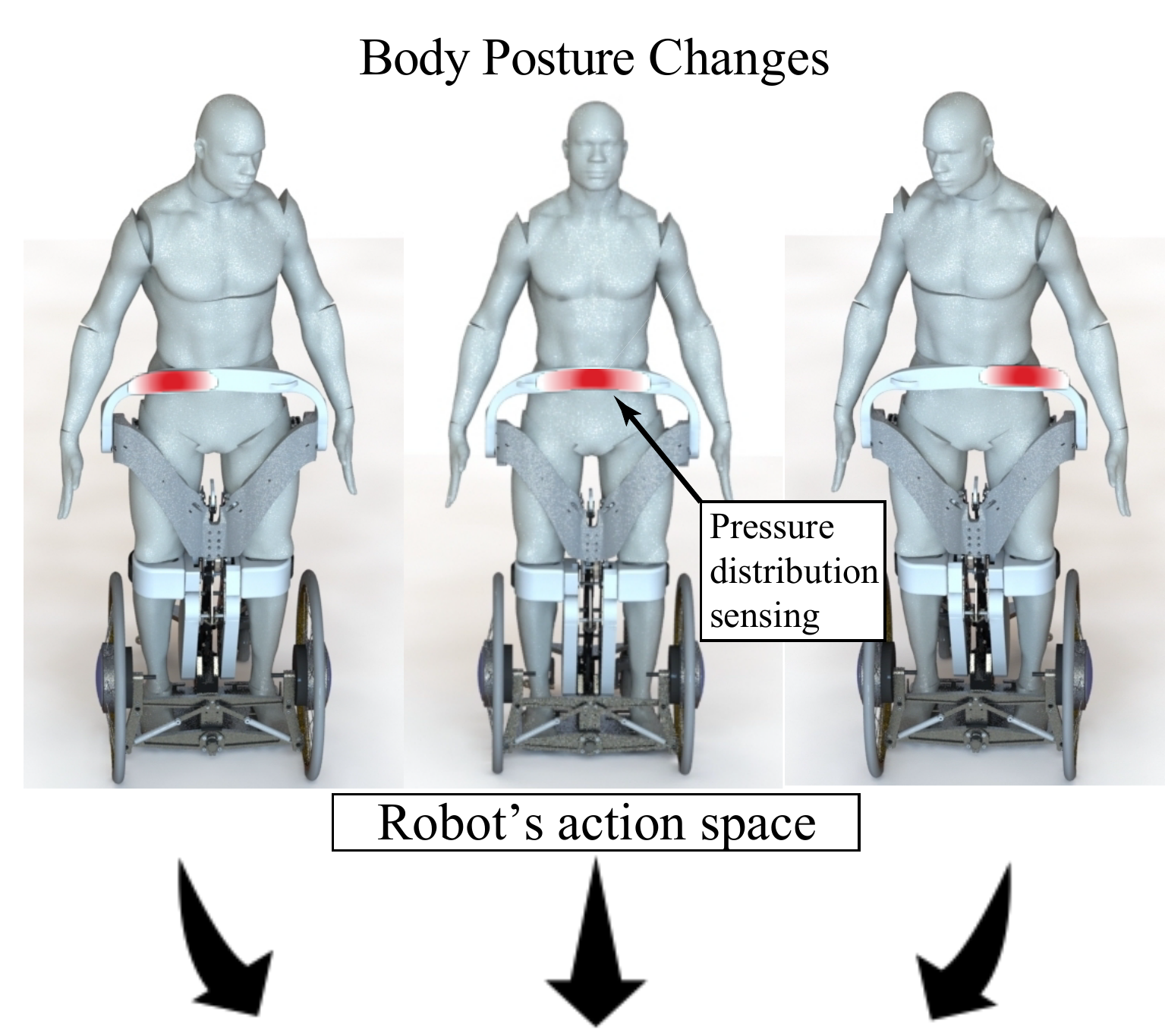}
      	\caption{Diagram of the proposed embodied torso based control in the standing mobility vehicle Qolo, for lower-body impaired people, depicting the pressure distribution in red based on the upper body motions from gaze tracking of the desired motion direction.
      	\label{fig:Qolo}}
   \end{figure}

Improving the way we interact with mobility devices through a human-machine interface (HMI) is an important goal for reducing the mental load to the user and increase the acceptability and usability of the device itself.
Qolo depicted in Fig. \ref{fig:Qolo} \cite{PaezGranados2018}, is a solution for standing mobility for lower-limb impaired people with a light-weighted system that combines a passive exoskeleton and powered wheeled base. This device allows a passive transition from sitting to standing or standing to sitting postures by the user voluntarily moving their centre of gravity, i.e., without external energy source required. Different models have been proposed for: lower-limbs impairment \cite{Eguchi2018}, lower-body impairment \cite{PaezGranados2018} and for children with lower-body impairment \cite{Sasaki2018}.
Hereof, the main question that we would like to address is: is there an intuitive and simple control interface for this type of personal mobility devices that would allow hands-free locomotion?

Currently, there are multiple approaches for solving standing mobility for wheelchair users, since 1975 when the Swiss company LEVO created the first modern standing wheelchair \cite{LEVO}, which provides both seating and standing postures in the static state; UPnRIDE (UPnRIDE Robotics Ltd., Yokne’am Illit, Israel) provides users with safe and functional mobility in a standing position in both indoors and outdoors with auto-balancing \cite{UPnRIDE}; Gyrolift (GYROLIFT, France) uses gyroscopes to allow the user to move in both sitting and quasi-standing posture \cite{GYROLIFT}; the Tek RMD (Matia Robotics, Inc., Salt Lake City, UT, USA) also offers standing and seating mobility \cite{TekRMD}. Nonetheless, all of them still rely on a joystick as main HMI unchanged from most powered wheelchairs or scooter-like handles. 
Undoubtedly, joysticks are one of the simplest and effective real-life solution HMI based on hand coordination control \cite{Marino1999}.  However, it limits the user to one limb and requires attention to the hand-eye coordination. 


Some recent work has proposed a shoulder-motion based HMI solution for cervical level spinal cord injury (SCI) through a set of inertia measuring unit (IMU) sensors \cite{Thorp2016}, however, this HMI would work only for the target users with little remaining motion capabilities on their upper-body. Equally, for severe mobility impairment (cervical injury level) brain-machine interfaces have been studied as shown in \cite{Carlson2013}. 
The closest solution to our proposed interface was presented in \cite{wakita2012riding}, which determined the operational intention by detecting the change in postures of a rider using pressure sensor sheets on a backrest, however, no detailed control or evaluation was developed, moreover, standing mobility was not considered.
A previous work of our group presented standing control by two potentiometers beside the user's hipbone connected by a link that allowed upper-body postures to be used as control inputs through the differential tilting angles \cite{Eguchi2018} . Nonetheless, this HMI requires free unsupported upper-body motions. 
The well-known two-wheeled self-balancing powered electric vehicle 'Segway' is very suitable for standing mobility, but its centre of mass (COM) balance control would not work for SCI people because it requires lower-body control \cite{Nguyen2004}.
In general, an absolute solution is not feasible for the wide differences in mobility devices and end-users' remaining motor control. e.g., eye movement tracking, head array, or sip-puff based controllers are current HMI for specific populations (a detailed review can be found in \cite{Simpson2005}). Therefore, we have focused on the control assistance of a user interface for wheelchairs or standing mobility devices for upper-body able users.

In this work, we propose a new type of control interface with natural body motions of the torso, which agrees with the idea of gaze and body movement introduced in \cite{Hollands2002}, depicted in Fig. \ref{fig:Qolo}. The space matching between body postures and robot control is done in a 2D space motion mapping, a preliminary idea introduced in \cite{chen2019}. The results show that users are capable of controlling the robot through it and achieve good performance similar to a joystick, although, yet to overcome it. 
The contributions of this work are: first, a general design of an interface for hands-free navigation in standing mobility vehicles, or person carrier robots (a category described in \cite{ISO15066}).
Second, a formal and generalized control principle with the corresponding algorithms (user intention recognition and user personalizing) based on upper-body movements. Third, an evaluation with a baseline interface showing its potential to be a daily life solution. Finally, evaluation and analysis of the muscle activity of healthy participants for inferring the level of motor control required for its usage.

The paper continues as follows: section \ref{sec_method} presents the overall design of the proposed control interface, and the natural body motion control algorithm. Section \ref{sec_sys} presents the system overview of the hardware. Sections \ref{sec_eval} and \ref{sec_res} present two experimental evaluation and results in terms of muscle activity, task completion time, and user perceptions. Section \ref{sec_con} concludes this work and addresses future work.

\section{METHODOLOGY}\label{sec_method}
\subsection{Human Machine Interface Design}
With the goal of achieving hands-free locomotion while minimizing the effort by the intended users, we present a methodology of embodied control based on the idea of that visual tracking during the gait dictates the body posture corrections, i.e., your body posture changes during walking based on the intended motion direction \cite{Hollands2002}.
With this hypothesis in mind, we designed an ergonomic support bar for upright posture in front of the user's lower-trunk, which embeds a matrix of pressure sensors on its surface, so that, the robot could sense slight upper-body motions and the center of pressure (COP).

\begin{figure}[http]
    \begin{center}
    \includegraphics[width=8.6cm]{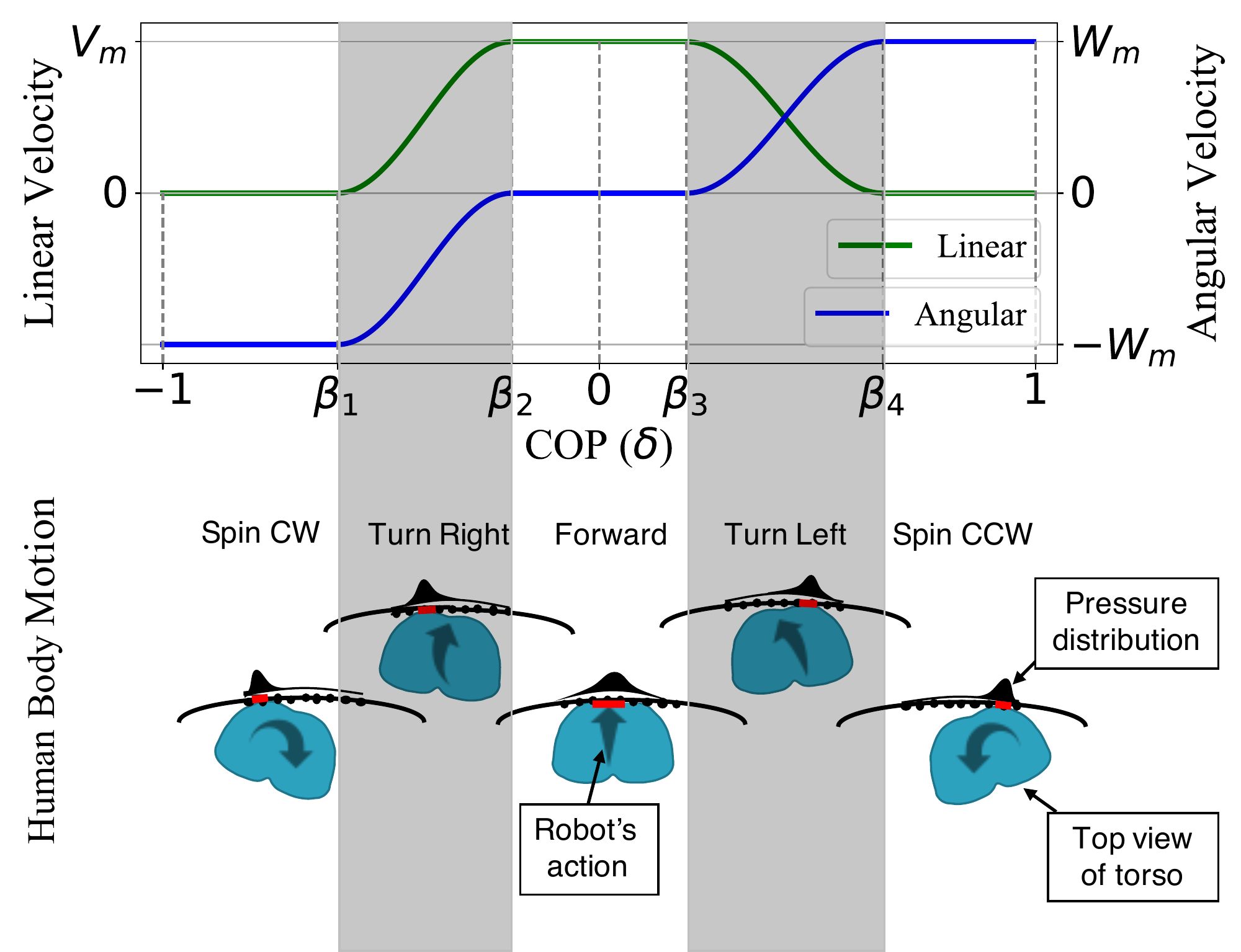}
    \caption{On the upper side: Expected Linear/Angular Velocity vs. Center of Pressure (COP); On the lower side: Expected posture-action, depicting the assumed motion of the user in blue, and the user intended directional motion output is represented by the arrows.}
    \label{oc}
    \end{center}
\end{figure}

The expected posture-action relationship is shown in Fig. \ref{oc}. Where movement control of the robot has been designed as a continuous function of the pressure magnitude and the COP.
This design intends to match the natural gaze tracking of the direction of motion while exploiting the user's upper-body residual motion capabilities. Although, the mapping between the human body and the established 2D space of the HMI implies an inverse kinematic problem with infinite possible human-body configurations in the muscle control space, we expect the users to learn one out of them as their optimal solution given their own body capacity.

Following the proposed function in Fig. \ref{oc}, there is a continuous action from the user's body motion on left to right, which maps a smooth response in the robot control space from spinning clockwise (CW) to spinning counter clockwise (CCW). 
On the other hand, backward motion is defined separately, a reason being that the backward movement is relatively dangerous, thus, it should be triggered by a conscious independent action. In this case, we propose pressing one extreme of the sensing array with a thumb and operate with the body or using thumbs on both extremes to operate. This action was derived from the user's natural behavior of holding the support bar to keep a sense of safety when looking backward.

\begin{figure}[!t]
      \centering
  		\subfigure[]{\includegraphics[width=2.5cm]{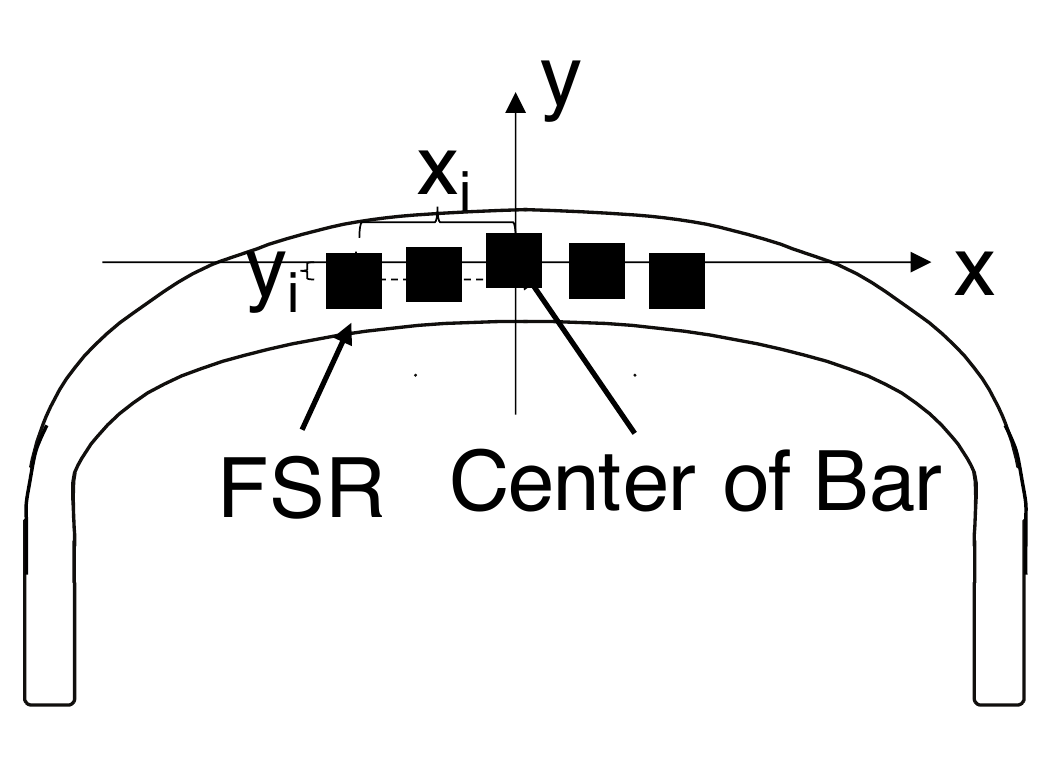}
		\label{fig:a}}
		\hfil
		\subfigure[]{\includegraphics[width=2.5cm]{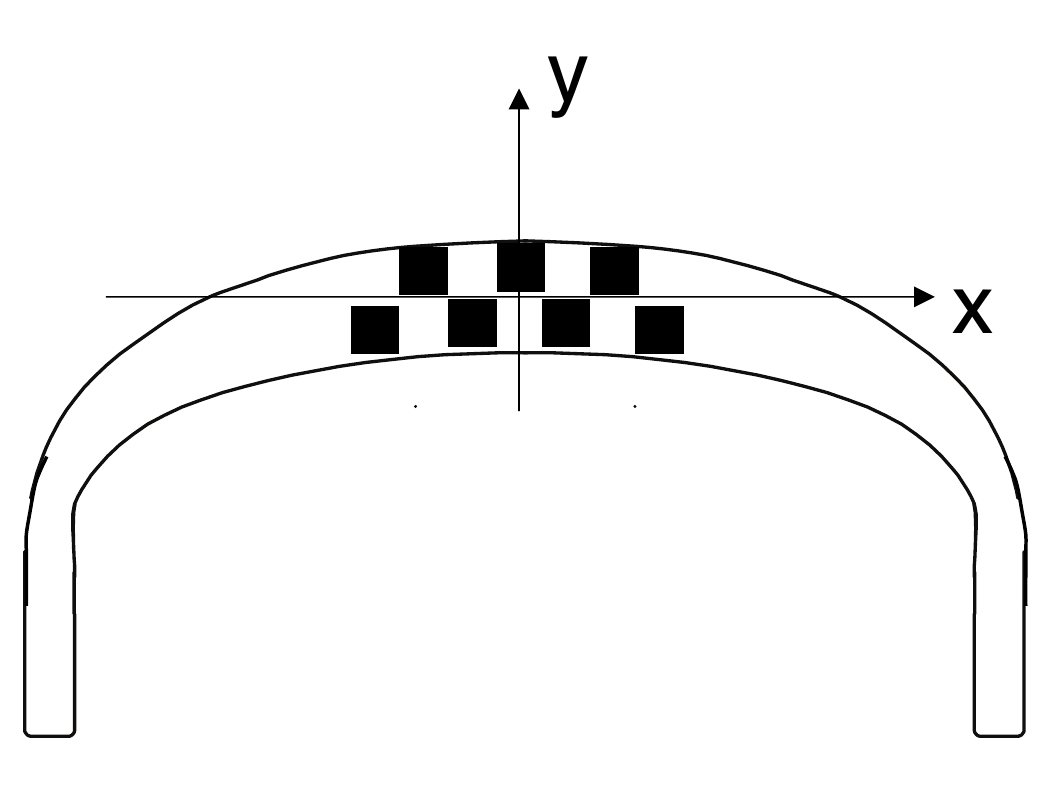}%
		\label{fig:b}}
		\hfil
		\subfigure[]{\includegraphics[width=2.5cm]{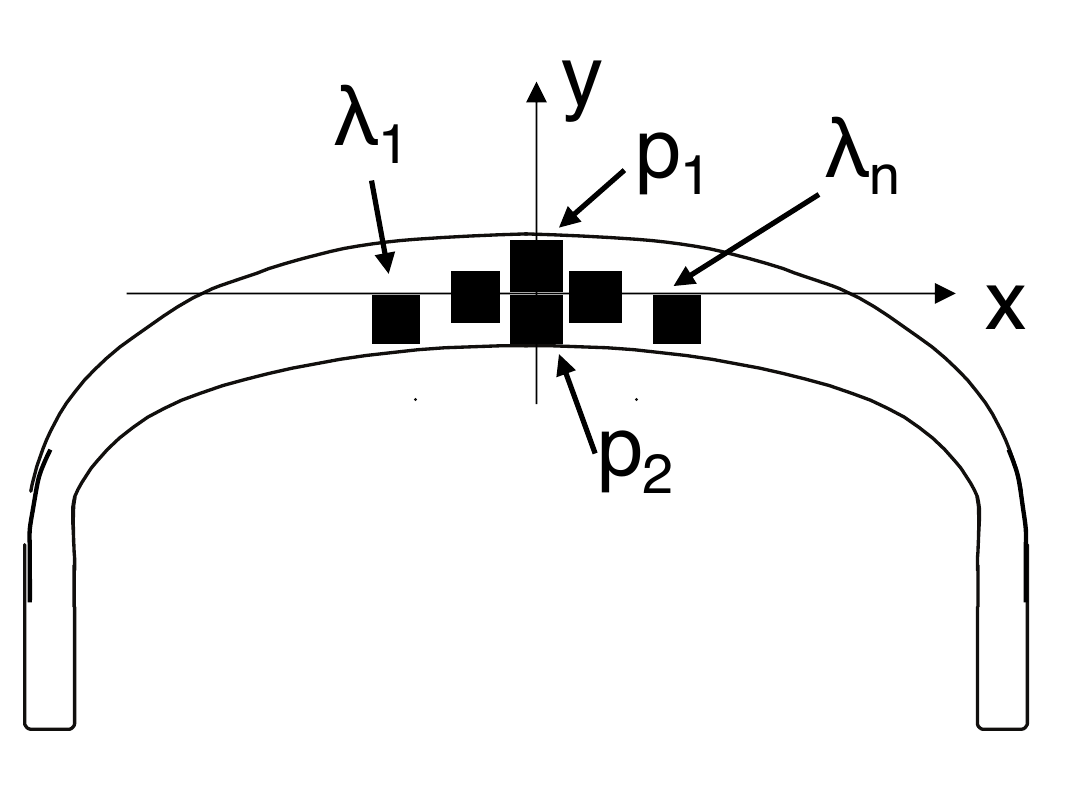}%
		\label{fig:c}}
      \caption{Distribution examples of pressure sensing matrices, with FSR depicted in black squares. In the current prototype design (a) was selected, however, any configuration would directly work with the proposed algorithms.
      \label{DIS} }
   \end{figure}

The design of the torso bar is an ergonomic ellipse, whose inner face is distributed on a 3D surface \cite{PaezGranados2018}. However, we only consider its projection over a 2D surface for designing the overall pressure distribution sensing. 
To lower the cost, we consider a simple FSR unit to form a sensing matrix. In this case, the distribution of FSR units could highly affect the detecting performance due to human-bar contacting conditions. Therefore, the design of the FSR distribution should be well considered.
A few example designs are shown in  Fig. \ref{DIS}. The origin of the coordinate is located at the center of torso bar, black squares denote sensors contact points. The location of $i$th sensor in coordinate is denoted as $(x_i, y_i)$. In Fig. \ref{fig:a}, FSR array aligns along the center curve the torso bar inner surface, which is the simplest case. Fig. \ref{fig:b} is designed to use more sensors to provide a higher precision of detecting in $x$ axis, Fig. \ref{fig:c} is designed to use more sensors in specific locations to compensate for the possible difficulties in detecting pressure distributions, e.g., due to the abdominal muscle and rib cage height and stiffness differences.
For higher detecting precision purposes the design of distribution of sensors could be replaced by a high precision pressure sensing matrix, with the drawback of increasing the cost of the system.

\subsection{Body Movement to Velocity Mapping}
The current formulation intends to generalize the user intend inference  for any number of discrete measuring points over a set space of possible contacts, so that, we propose a generic system that maps the expected human body motions directly to the robot's control space $\xi = [v, \omega]'$, where $v$ denotes the linear velocity, and $\omega$ the angular velocity, both in the local frame of the robot. 
The HMI input space $u=[ \delta, P]$ is defined by the center of pressure ($ \delta$), and the maximum pressure ($P$) measured over the sensors' array as follows:
\begin{eqnarray}
\delta =\dfrac {\sum ^{n}_{i=1} \alpha_i \overline{\lambda }_{i}s_{i}}{\sum ^{n}_{i=1} \overline{\lambda }_{i}}
\label{COP}\\
 P =max\left( \overline {\lambda }_{i}\right) 
\label{Lambda}
\end{eqnarray}

where $\alpha_i$ denotes a weight to the defined sensor column $i$, for a weighted sum based on the system calibration. $\overline{\lambda_i}$ defines the mean of sensor readings for column $i$, as $\sum_{j=1}^{m} p_{j} / m$. This can be encapsulated in the matrix $F=\left[ \lambda _{1},\lambda _{2},\ldots ,\lambda _{n}\right]$ with column vectors $\lambda _{i}=\left[ p_{1},p_{2},\ldots ,p_{m}\right] ^{-1}$, corresponding to each sensor $p_{j}$ with equal $x_i$ for $i \in \mathbb{R}^{n} $ as in Fig. \ref{DIS}. Finally, the term $S=[ s_{1},s_{2},\ldots ,s_{n}]$ denotes the absolute location of the column of sensors $\lambda_i$ as:
 \begin{equation}
 s_{i}=\dfrac {x_{i}}{\left\| x_{n}-x_{1}\right\| /2},
  \label{sensor_pos}
 \end{equation}
within the interval $-1 \leq s_i \leq 1 $, for any sensor location in the $x$ axis $x_1 \leq x_i \leq x_n$.

Then a proportional control defines the mapping to speed magnitude of the robot $\xi_m \in [v_m, \omega_m]'$ from the maximum input pressure given by the user $P$, so that, $\xi_m = K P$, for a gain matrix $K = [k1, k2]'$, \(v_m\) the linear component, and \(\omega_{m}\) the angular component constrained the robot's limits ($v_{max},  \omega_{max}$) $0 \leq v_m \leq v_{max}$ and $0 \leq \omega_m \leq \omega_{max}$ by the gains, $k1 \leq v_{max} / P_{max}$ and $k2 \leq \omega_{max} / P_{max}$, with the sensor maximum value $P_{max}$ as a constant obtained from an experimental measurement.

Subsequently, we map the input motion space to the robot's control space, by a piece-wise function varying from linear to angular velocity in function of the COP ($\delta$) as follows:
\begin{equation}
v(\delta)=\left\{
\begin{aligned}
&0, \quad \quad \quad\quad \quad  \quad  \quad \quad  \quad \quad  \quad  \quad -1\leq \delta < \beta_1 \\
&\frac{v_{m}}{2} + \frac{v_{m}}{2}\sin(\frac{\pi}{\beta_1-\beta_2}(\delta-\beta_1) + \frac{\pi}{2}),  \\
&\quad \quad  \quad  \quad  \quad  \quad \quad  \quad  \quad  \quad  \quad  \quad  \quad  \beta_1 \leq \delta < \beta_2 \\
&v_{m}, \quad \quad \quad \quad \quad \quad \quad \quad \quad \quad \beta_2 \leq \delta < \beta_3 \\
&\frac{v_{m}}{2} + \frac{v_{m}}{2}\sin(\frac{\pi}{\beta_4-\beta_3}(\delta-\beta_3) + \frac{\pi}{2}), \\
&\quad \quad \quad  \quad  \quad  \quad  \quad  \quad  \quad  \quad  \quad  \quad  \quad \beta_3 \leq \delta < \beta_4 \\
&0, \quad  \quad \quad  \quad  \quad \quad \quad  \quad  \quad  \quad  \quad \quad \beta_4 \leq \delta \leq 1 \\
\end{aligned}
\right.
\end{equation}

\begin{equation}
w(\delta)=\left\{
\begin{aligned}
&-\omega_{m}, \quad \quad \quad \quad \quad \quad \quad \quad \quad -1 \leq \delta < \beta_1 \\
&\frac{-\omega_{m}}{2} - \frac{\omega_{m}}{2}\sin(\frac{\pi}{\beta_1-\beta_2}(\delta-\beta_1) + \frac{\pi}{2}),  \\
&\quad \quad  \quad  \quad  \quad  \quad \quad  \quad  \quad  \quad  \quad  \quad  \quad  \beta_1 \leq \delta < \beta_2 \\
&0, \quad \quad \quad\quad \quad  \quad  \quad \quad  \quad \quad  \quad  \quad \beta_2\leq \delta < \beta_3 \\
&\frac{\omega_{m}}{2} + \frac{\omega_{m}}{2}\sin(\frac{\pi}{\beta_4-\beta_3}(\delta-\beta_4) + \frac{\pi}{2}), \\
&\quad \quad \quad  \quad  \quad  \quad  \quad  \quad  \quad  \quad  \quad  \quad  \quad \beta_3 \leq \delta < \beta_4 \\
&\omega_{m}, \quad \quad \quad \quad \quad \quad \quad \quad \quad \quad \quad \beta_4 \leq \delta < 1 \\
\end{aligned}
\right.
\end{equation}
where \(\beta_1,\beta_2,\beta_3,\beta_4\) are classification points for COP $\delta$, a more intuitive mapping is shown in Fig. \ref{oc}.

\subsection{Calibration for Body Motion Input}
As people come in different shapes and sizes, moreover, people have different proprioceptive control of their bodies, we designed the controller with a set of classification points \(\beta_1, \beta_2, \beta_3, \beta_4\) for adapting to each user, thus, allowing body-specific control. 

Here, we propose a personalizing algorithm that calibrates once per subject and could be called by the user if wanted by an independent trigger from an onboard sensor attached on the exterior side of torso bar. Generally these devices are single-user oriented, thus, calibration might be wanted only if the user body ability changes.

A first pre-calibration aims to balance human intentions and pressure reading, which is expressed in 2 folds, minimum press intention corresponding to a zero offset, and maximum press intention corresponding to $P_{max}$. The minimum press intention indicates a neutral posture or relaxed state, where the user would like to stop all motions. While, the maximum press intention indicates a posture where the user would like to achieve the highest velocity. Herewith, the pressure distribution under different body shapes is expected to be as similar as possible to the designed one depicted in Fig. \ref{oc}. 
The minimum press intention calibration simply collects a certain amount of reading values of each sensor under neutral or relaxed posture of the user, getting the mean of each sensor as zero offset. On the other hand, the maximum press intention is obtained by getting $\overline{\lambda_i}$ under human's maximum press intention, thus, obtaining the corresponding weights $\alpha_i$ to satisfy the equality:
\begin{equation}
P_{max} = \alpha_1\overline{\lambda_1} = \alpha_2\overline{\lambda_2} = ... = \alpha_n\overline{\lambda_n}
\label{max}
\end{equation}

Calibration procedure includes two steps as summarized in algorithm 1, the first step is to collect the sensor readings, through a known process: we ask the user to switch from "Spin CW" to "Spin CCW" posture with maximum press intention, then move in the reverse order for a set of ten transitions staying at each posture 5 seconds (indicated by an LED on board). 
In total $N_s$ sets of readings are recorded, the top ten percent readings of each single sensor will be summed up and get the average value: $\overline{\lambda_{1v}}, \overline{\lambda_{2v}}..., \overline{\lambda_{nv}}$, then based on Equation \ref{max}, we could get the value of \(\alpha_1, \alpha_2, ..., \alpha_n\), which will be taken as coefficients for the user.

To generate the personalized control points $\beta_1, \beta_2, \beta_3, \beta_4$, we calculate the COP \(\delta\) for each posture as \(\delta_1\), \(\delta_2\), \(\delta_3\), \(\delta_4\), \(\delta_5\), then the calibrated classification points are obtained as: \(\beta_1 = (\delta_1 + \delta_2)/2, \beta_2 = (\delta_2 + \delta_3)/2, \beta_3 = (\delta_3 + \delta_4)/2, \beta_4 = (\delta_4 + \delta_5)/2\).

\begin{algorithm}[h]
  \caption{Calibration Algorithm}
  \begin{algorithmic}[1]
\Procedure{Calibration}{$P_{max}$, $N_s$, $\overline {\Lambda_1}$,$ \overline {\Lambda_2}$, ... ,$ \overline {\Lambda_n}$}
	\State $i$ $\leftarrow$ 1;
      \While {$i \leq n$}
    \State $ \overline{\lambda_{iv}}$ $\leftarrow$ mean(largest(0.1length($ \overline {\Lambda_i}$), $ \overline {\Lambda_i}$));
    \State $\alpha_i$ $\leftarrow$ $P_{max}$ / $ \overline{\lambda_{iv}}$;
     \EndWhile

	\State $i$ $\leftarrow$ 1; $j$ $\leftarrow$ 1;
      \While {$j \leq N_s$}
            \State Getting the COP $\delta[j]$ as Equation (1)
      \EndWhile
	\State $i$ $\leftarrow$ 1;
      \While {$i \leq 5$}
    \State $\delta_i$ $\leftarrow$ (mean($\delta[(N_s(i-1)/10)+1:N_si/10]$) + mean($\delta[(N_s(10-i)/10)+1:N_s(11-i)/10]$)) / 2;
      \EndWhile
	\State i $\leftarrow$ 1;
      \While {$i \leq 4$}
    \State {$\beta_i$} $\leftarrow$ {$(\delta_i + \delta_{i+1})/2$};
    \EndWhile

\Return {$\alpha_1, \alpha_2, ..., \alpha_n, \beta_1, \beta_2, \beta_3, \beta_4$};

\EndProcedure
  \end{algorithmic}
\end{algorithm}

\section{System Overview}\label{sec_sys}
\subsection{Pressure sensing system}\label{sec_bar}
\begin{figure}[!t]
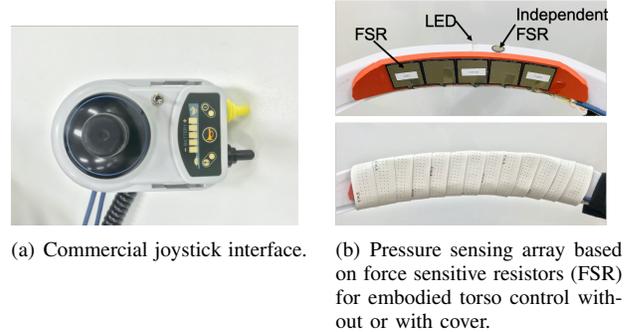

      \centering
  		\subfigure[Commercial joystick interface.]{\includegraphics[width=1.5in]{joystick.pdf}
		\label{fig:joystick}}
		\hfil
		\subfigure[Pressure sensing array based on force sensitive resistors (FSR) for embodied torso control without or with cover.]{\includegraphics[width=1.5in]{structure_double.pdf}%
		\label{fig:FSR}}
      \caption{Pictures of the evaluated motion control interfaces.
      \label{fig:interfaces} }
   \end{figure}

For the current design we have opted to minimize the number of pressure sensors, herewith, selecting 5 sensors for classifying user's upright postures, which is considered as a minimum number of sensors to control the robot motion. The size of sensor is 44$\times$44 mm, The distance between each sensor is 5mm, so that the total covered length is 220 mm, which could adapt to a wide number of user's waistline. A detailed view of the pressure sensing array is shown in Fig. \ref{fig:FSR}.

\subsection{Embedded control system}
\begin{figure}[t]
    \begin{center}
        \includegraphics[width=8cm]{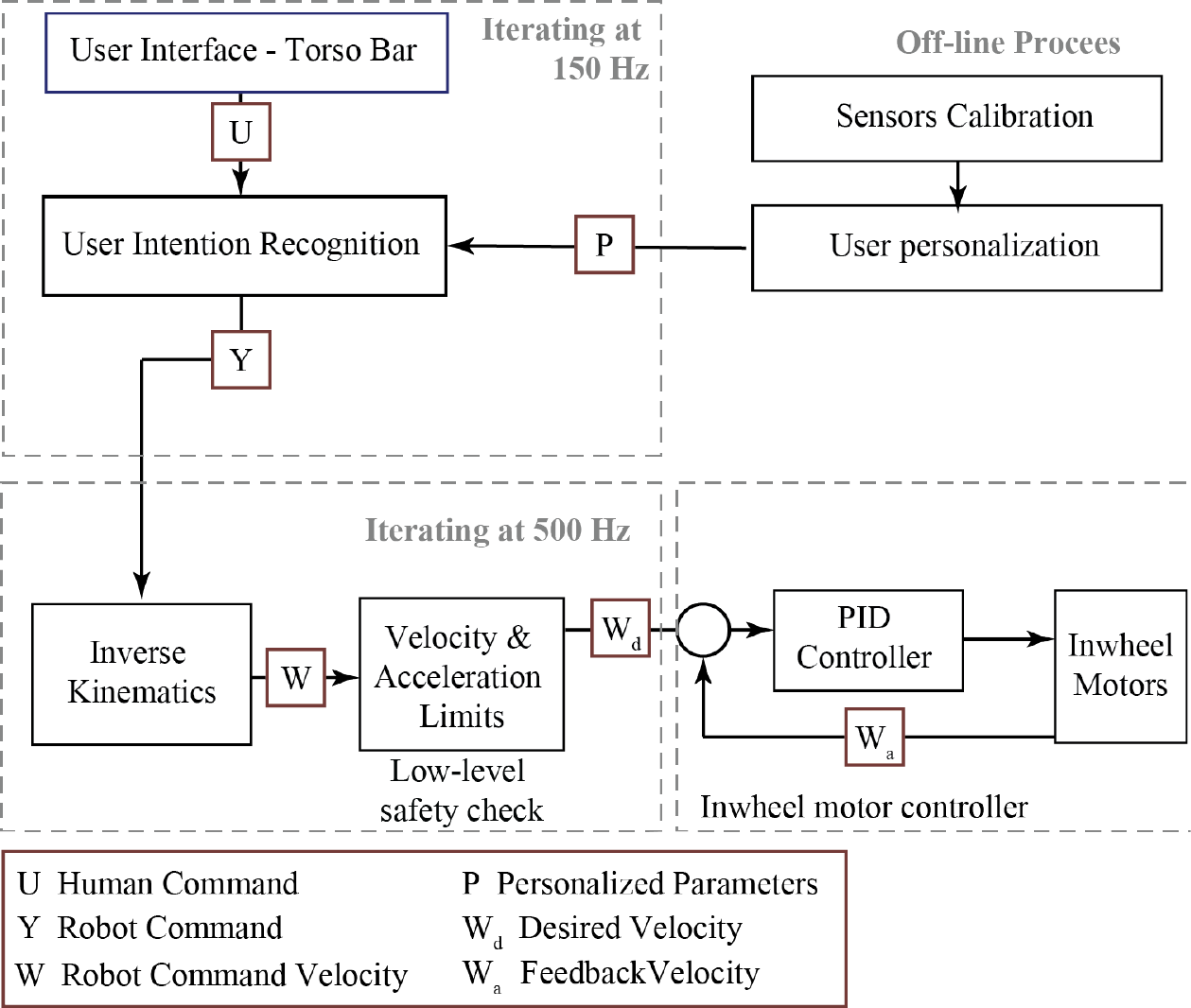}
        \caption{Overall control system architecture, considering the low-level velocity controller for the mobile base of the robot, and the higher level user intention recognition coupling. }
        \label{system_overview}
    \end{center}
\end{figure}

An UP-Squared (AAEON Technology Inc.) was selected as single-board computer combined with a High-Precision AD/DA Board (Waveshare) for reading the sensors input and communicate through analog channels to the actuators (YAMAHA Motor Corp., Iwata, Japan). The overall control architecture is presented in Fig. \ref{system_overview}, where the higher level currently operates at $150 Hz$ including sensing and user intention recognition algorithm. While the lower-level controller deals with kinematics and constraints for the robot motion at $500Hz$. Finally, the inner in-wheel motor control loop operates with a PID controller in the proprietary motor drive.

\section{EXPERIMENTAL EVALUATION}\label{sec_eval}
A two-part experiment was designed for assessing the proposed control interface. The first part, focused on a biomechanical viewpoint aiming to understand the human body motion by measuring the muscle groups required to drive in standing posture with the proposed torso HMI. Herewith, answering the question: what motor control ability requires a potential user?
The second part aimed to observe the performance of the overall system, by means of comparing it to a standard joystick control, here, focusing on quantitative and qualitative analysis of the user perceptions of usability \cite{brooke1996quick} and likeness \cite{Bartneck2009}.
A total of 14 participants joined the experiments, six participants joining the first group and eight in the second group.

   \begin{figure}[!t]
      \centering
  		\subfigure[EMG recording setting for training on the device control. ]{\includegraphics[width=7.0cm]{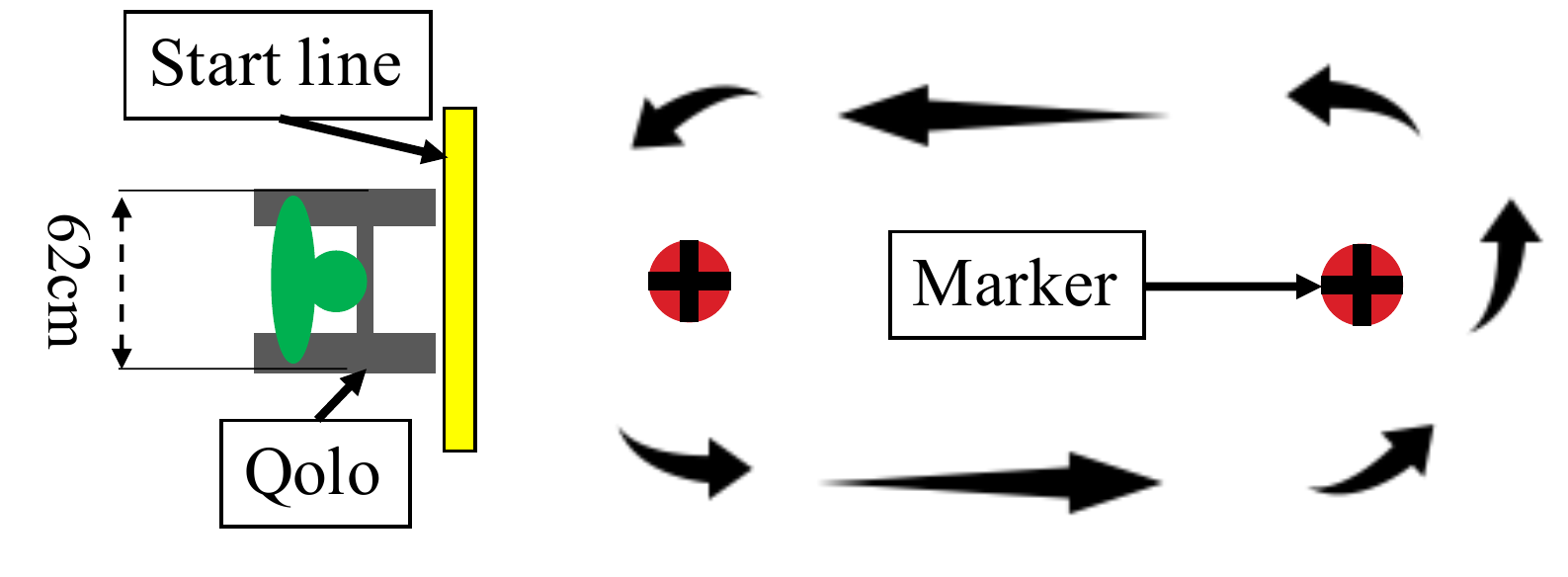}
		\label{fig:6p}}
		\hfil
		\subfigure[Motion control evaluation circuit.]{\includegraphics[width=7.0cm]{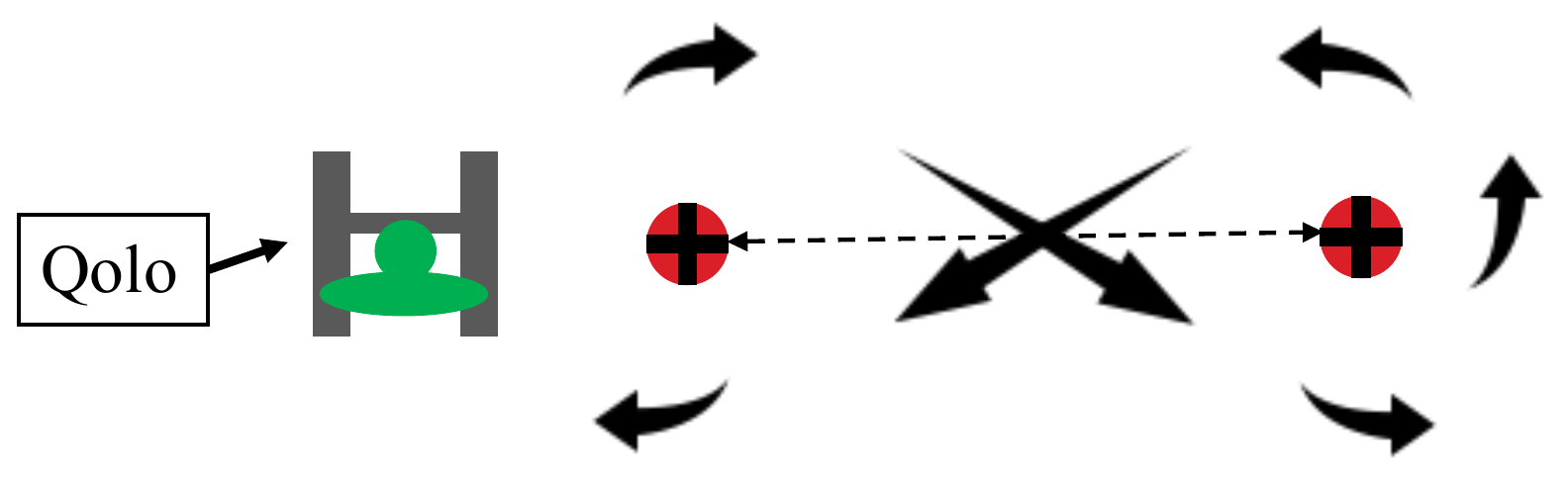}%
		\label{fig:8p}}
      \caption{Experimental setup: depicting with arrows the path set for subjects to follow during the test around the marked center points in left and right.
      \label{EE} }
   \end{figure} 
   
\subsection{Torso Muscles Control Analysis}
For the first set of experiments, a group of six subjects, all males (age $24.3 \pm 4.7 $ years old, height $173.8 \pm 5.8 $ cm, and weight $65.3 \pm 8.9 $ kg) joined the experiments. The general procedure was: 1- calibration and explanation of the torso interface, 2- training session, 3- evaluation.

The participants were instructed on the torso control, and then allowed to train over a lapse of $45$ minutes by moving in clockwise and counterclockwise circles around two markers set a part $5 m$ and then back to the start point as depicted in Fig. \ref{fig:6p}. This training was considered enough time for learning the HMI and controlling the robot for the task at hand in a natural way by all participants.
The evaluation consisted in controlling the standing vehicle Qolo for performing an 8-like shape as depicted in Fig. \ref{fig:8p} as continuously as possible during 3 circuits.

A set of electromyography (EMG) sensors (Trigno Lab, Delsys) were used to record the muscles activity in the upper body during the specified task. The EMG sensors were located at the erector spinae (ES) in lumbar (proximal) and thoracic (distal) region to observe the torso extension, at the rectus abdominis (RA) at level T6 (proximal) and T12 (distal);  external abdominal oblique (AO) in lumbar region; and at the pectoralis major (PM).
As well, a motion capture system (MX System, Vicon Motion Systems, Ltd.) in synchrony with the EMG system was used to record the motion trajectories with markers attached at the ankle, knee, pelvis, shoulders, elbows, and wrists.

\subsection{Comparison with a Baseline Control Interface} \label{sec_com}
In the second part of the evaluation, a group of eight subjects all males (height $172.2 \pm 5.8 $ cm, weight $71.5 \pm 8.2 $ kg, and body mass index (BMI) $24.2 \pm 2.6$ $kg/m^2$) joined the experiments (detailed characteristics on TABLE \ref{table:users}).
The evaluation of the designed interface consists of controlling Qolo to move in a circuit routine around two markers in an 8-like shape and then back to the start point as depicted in Fig. \ref{fig:8p}. The embodied interface as shown in Fig. \ref{fig:FSR} was compared to the standard commercial joystick interface in Fig. \ref{fig:joystick} as a baseline measurement. 

\begin{table}
\centering                
\begin{tabular}{c|c|c|c|c|c}    
\hline                
  & $P_1$ & $P_2$ & $P_3$ &$ P_4$ \\
\hline
Weight(kg) & 54 & 68 & 74 & 72 \\
\hline
Height(cm) & 170 & 174 & 172 & 173 \\
\hline
BMI($kg/m^2$) & 18.68 & 22.46 & 25.01 & 24.5 \\
\hline
Xiphisternum(cm) & 15.4 & 8.1 & 15.8 & 14.2 \\
\hline
& $P_5$ & $P_6$ & $P_7$ & $P_8$ \\
\hline
Weight(kg) & 75 & 79 & 80 & 70 \\
\hline
Height(cm) & 178 & 179 & 172 & 160 \\
\hline
BMI($kg/m^2$) & 23.67 & 24.65 & 27.04 & 27.34 \\
\hline
Xiphisternum(cm) & 14.6 & 15.3 & 13.4 & 4.9 \\
\hline
\end{tabular}             
\caption{Healthy participants information: Xiphisternum indicates the distance between the lowest point of the sternum and highest point of torso bar}
\label{table:users}
\end{table} 
All users were novice to wheelchairs and performed the evaluation in a set of 2 sessions for each control interface with a short break of 15 minutes in between.
Prior to the experiment, the users received an explanation of the control interface and the calibration algorithm was executed for each of them. As well, a short practice was allowed prior to the test recordings, so that, they could understand their body motions for achieving the basic inputs as depicted in Fig. \ref{oc}.

All participants tested both control interfaces, thus, for counterbalancing half of them followed $joystick \rightarrow torso$ interface, and the other half $torso \rightarrow joystick$.
In order to observe their first impressions on the torso interface and observe how effectively could be controlled by a novice, we divided each evaluation into 2 sessions, each of six continuous circuits (Fig. \ref{fig:8p}). 
Out of these tests, we evaluate the following metrics:
\begin{itemize}
    \item the task completion time $T = t_f - t_0$.
    \item the overall motion jerk $J = (1/T*N) \sum ^{N}_{i=1} ||\dot{\xi_i} ||$
    \item the user input fluency $F = (1/N) \sum ^{N}_{i=2} ( 1- |\Lambda_i -\Lambda_{i-1} |) $
\end{itemize}

Detailed information about the users in TABLE \ref{table:users} informs of possible trends in the results based on the fitting to the robot, specially, considering the pressure sensing location on the body through the Xiphisternum distance to the center of the pressure array. Only user $P8$ had significantly higher contact with the sensors in his body, however, results showed no significant variations in the usage capabilities for any subject.

   \begin{figure}[!t]
      \centering
  		\subfigure[Snapshots during EMG recording experiments.]{\includegraphics[width=8.0cm]{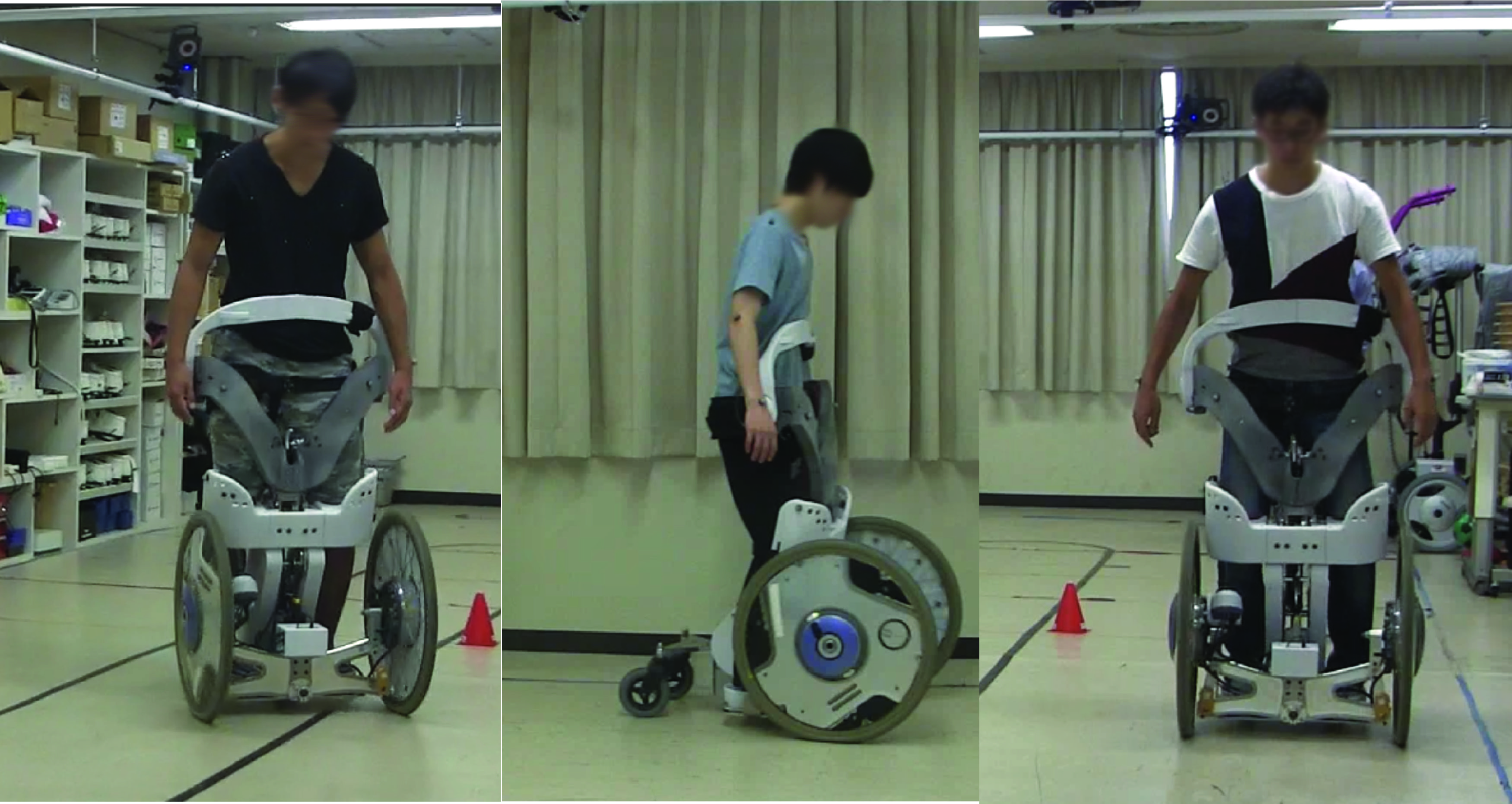}
		\label{fig:emg_subjects}}
		\hfil
		\subfigure[Motion sequence of a participant driving Qolo. ]{\includegraphics[width=8.0cm]{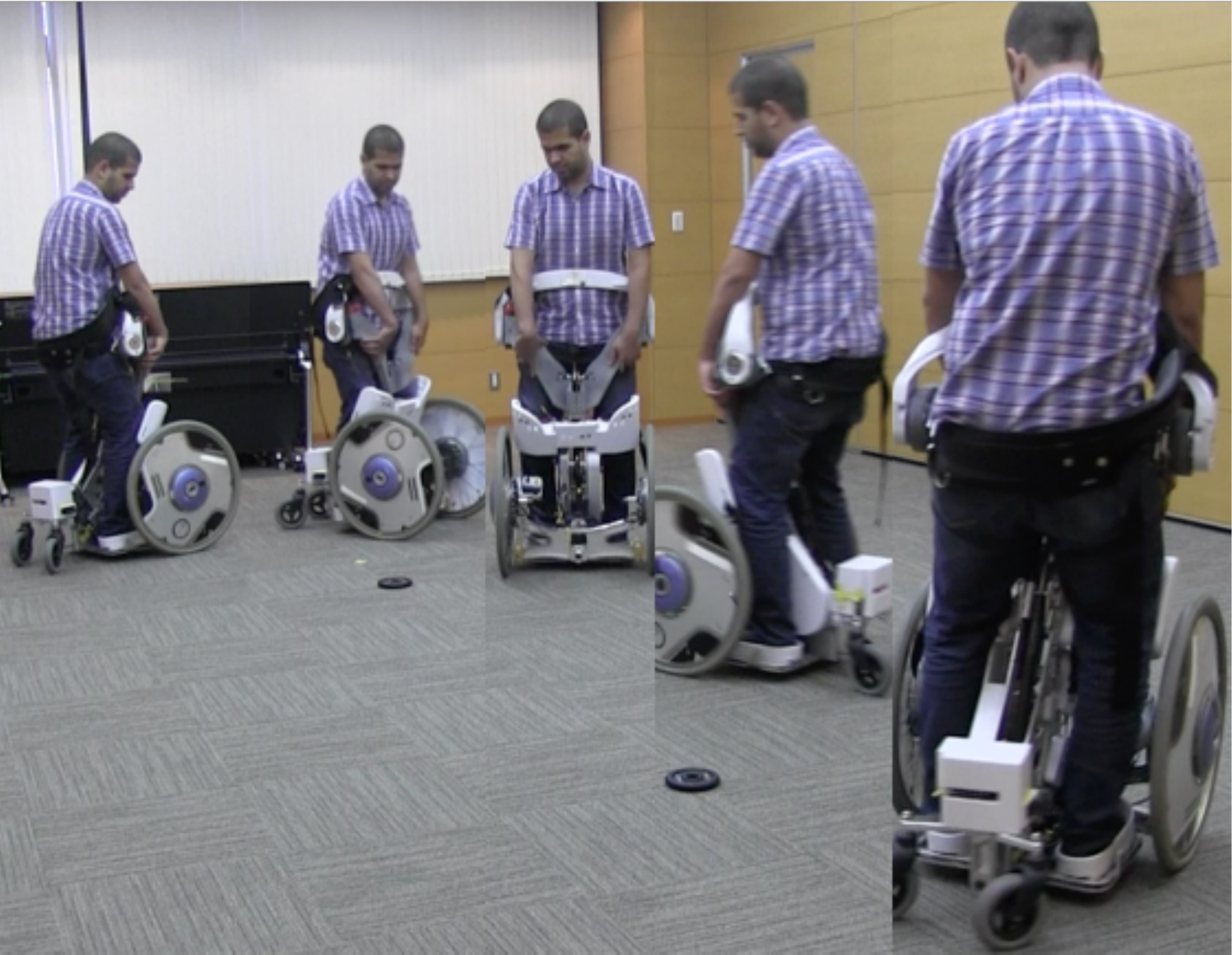}%
		\label{fig:p3}}
      \caption{Experimental snapshots of different subjects driving the robot Qolo with the proposed interface.
      \label{fig:Subjects} }
   \end{figure} 

\section{RESULTS AND ANALYSIS} \label{sec_res}

\subsection{Torso Muscles Control Analysis from EMG}
\begin{figure}[t]
        \centering
            \includegraphics[width=8.0cm]{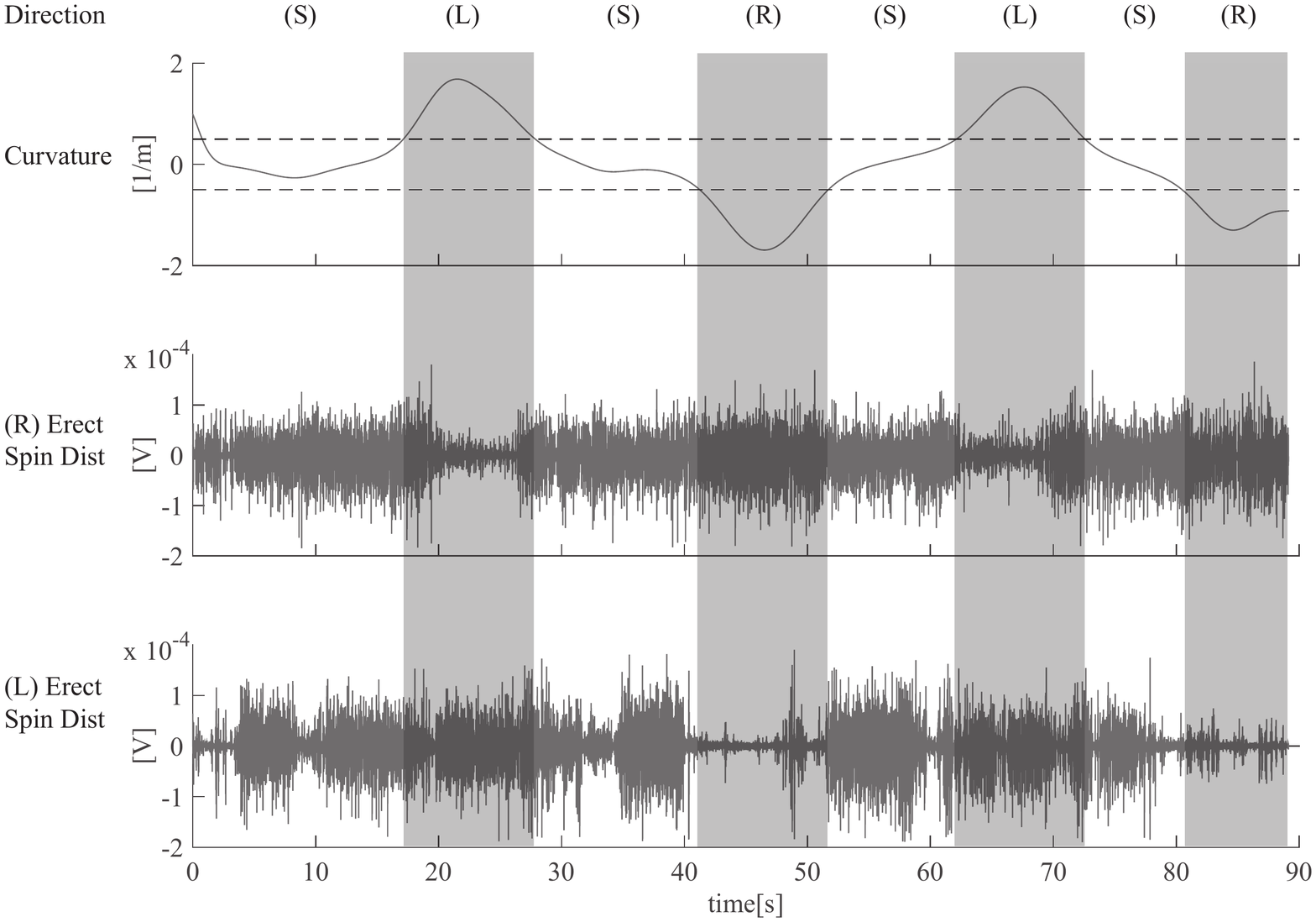}
        \caption{EMG data sample for Erector Spinae, divided in bars that  classify the motion type based on the robot's recorded motion, left turn (L), right turn (R), and straight (S).}
        \label{fig:EMG_sample}
    \end{figure}

Trunk muscles usage during driving control of the standing mobility vehicle through the proposed torso interface was analyzed with 6 healthy participants. A sample of the EMG recording for the selected trajectory is presented in Fig. \ref{fig:EMG_sample}. 
We took into account maneuvers for the last 2 circuits in an 8-like shape circuit as described in Fig. \ref{fig:6p} for comparing muscle activation.

\begin{figure}[t]
        \centering
            \includegraphics[width=8.0cm]{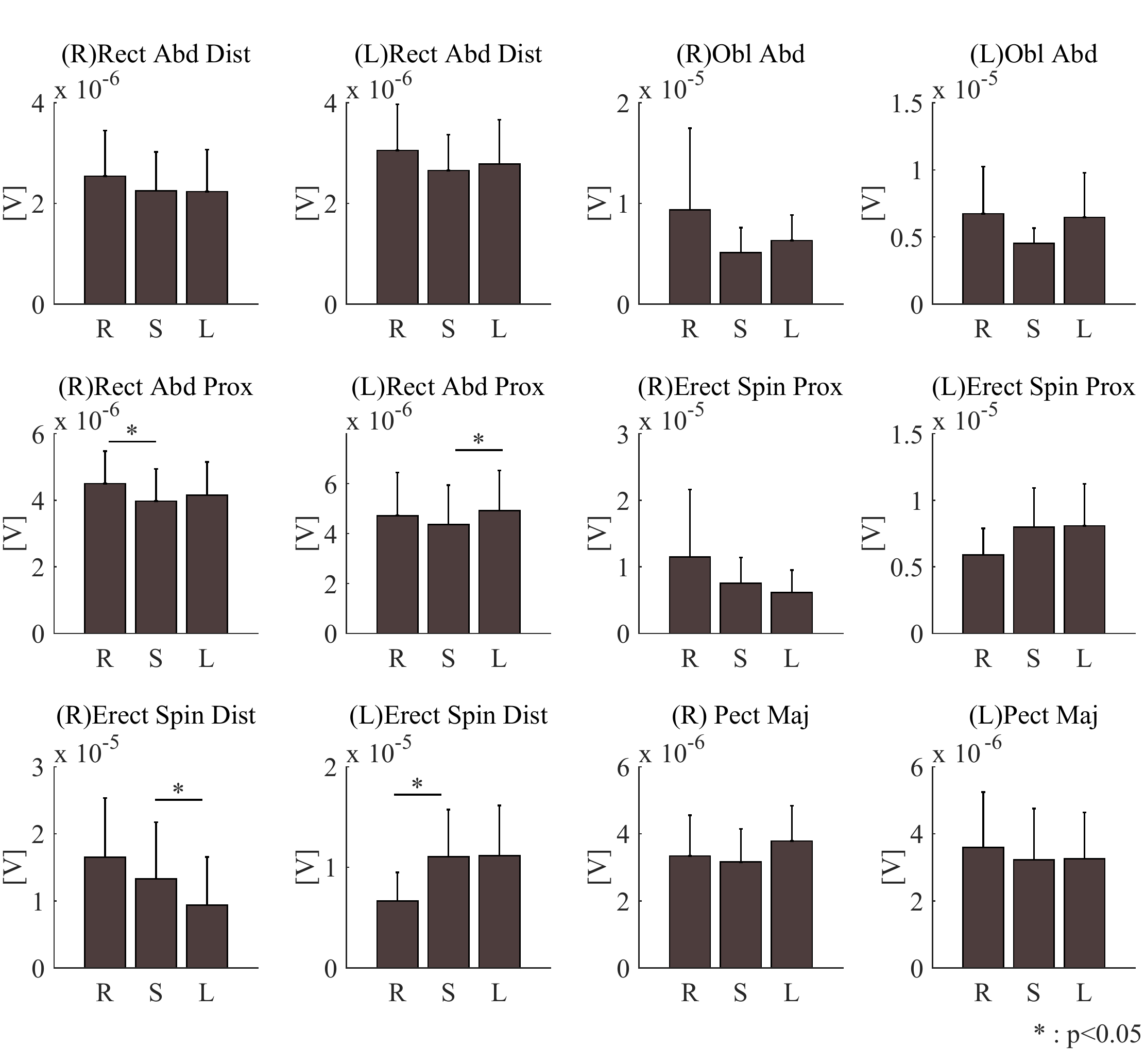}
        \caption{Results of EMG analysis of the 6 subjects for the torso test, showing the mean and standard deviation over each muscle group assessed per individual subject comparing between turning right (R), left (L) and moving straight (S).}
        \label{fig:EMG_results}
    \end{figure}

The results presented in Fig. \ref{fig:EMG_results}, highlight the sets of groups that are significantly active during lateral motion control compared to straight driving. These comparison between of turning maneuvers showed that right turns activation of right proximal rectus abdominis (RA) increased by $14 \%$ and the left distal erector spinae (ES) decreased by $36 \%$. 
In the same manner, when turning to left, left RA increased by $14 \%$ , and right ES decreased by $32 \%$. 
All other muscle groups measured showed no significant difference in the maneuvering of the robot.
These results, consistent among all subjects suggest that lumbar level ES and RA (high level) are the most relevant muscle groups involved in controlling the robot with the proposed HMI.

\subsection{Comparison with a Baseline Control Interface}
\begin{figure}[!t]
      \centering
		\subfigure[Velocity output in the robot during joystick control.]{\includegraphics[width=5.0cm]{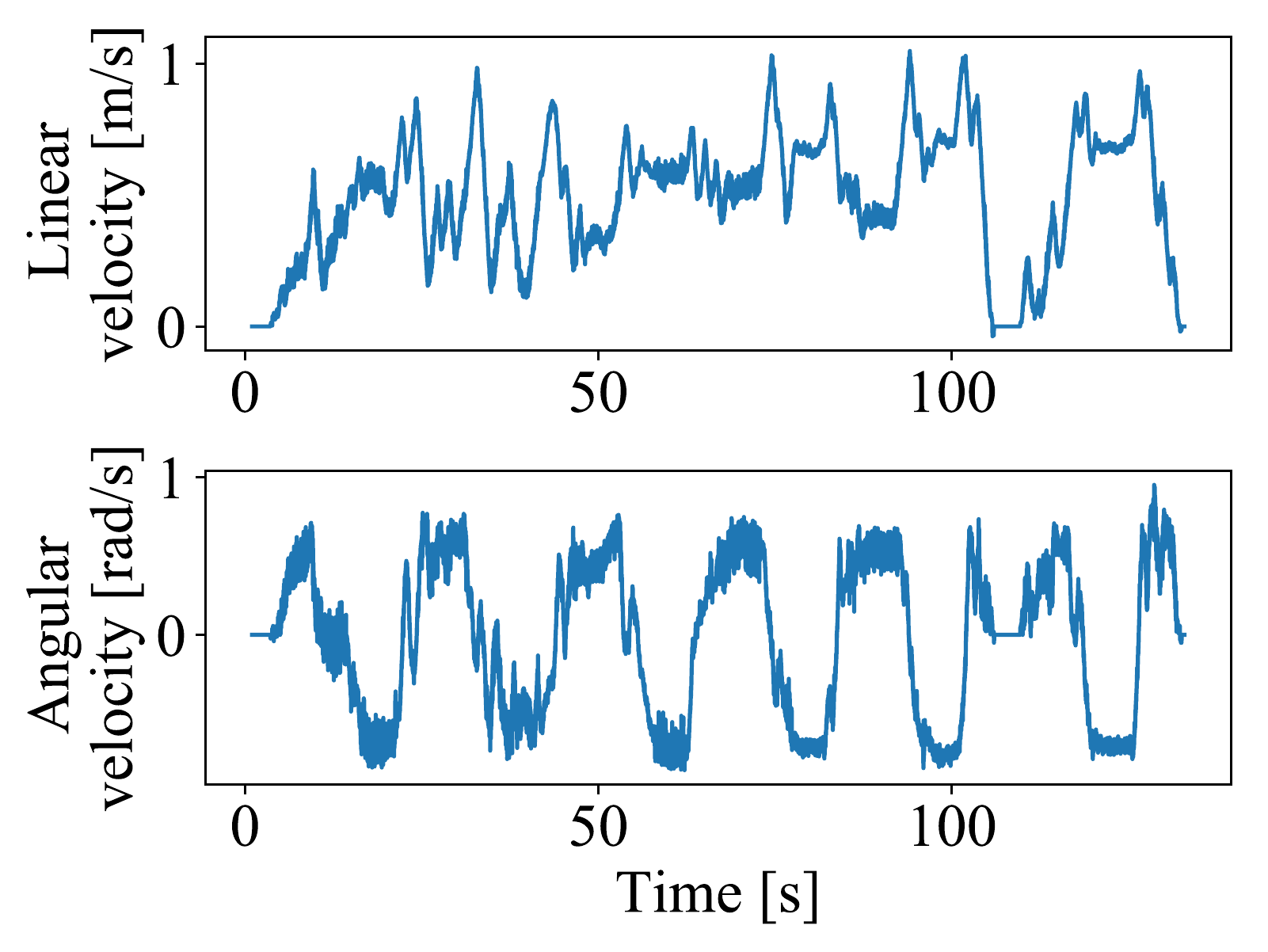}%
		\label{fig:vel_j}}
		\hfil 
		\subfigure[Velocity output in the robot during torso control.]{\includegraphics[width=5.0cm]{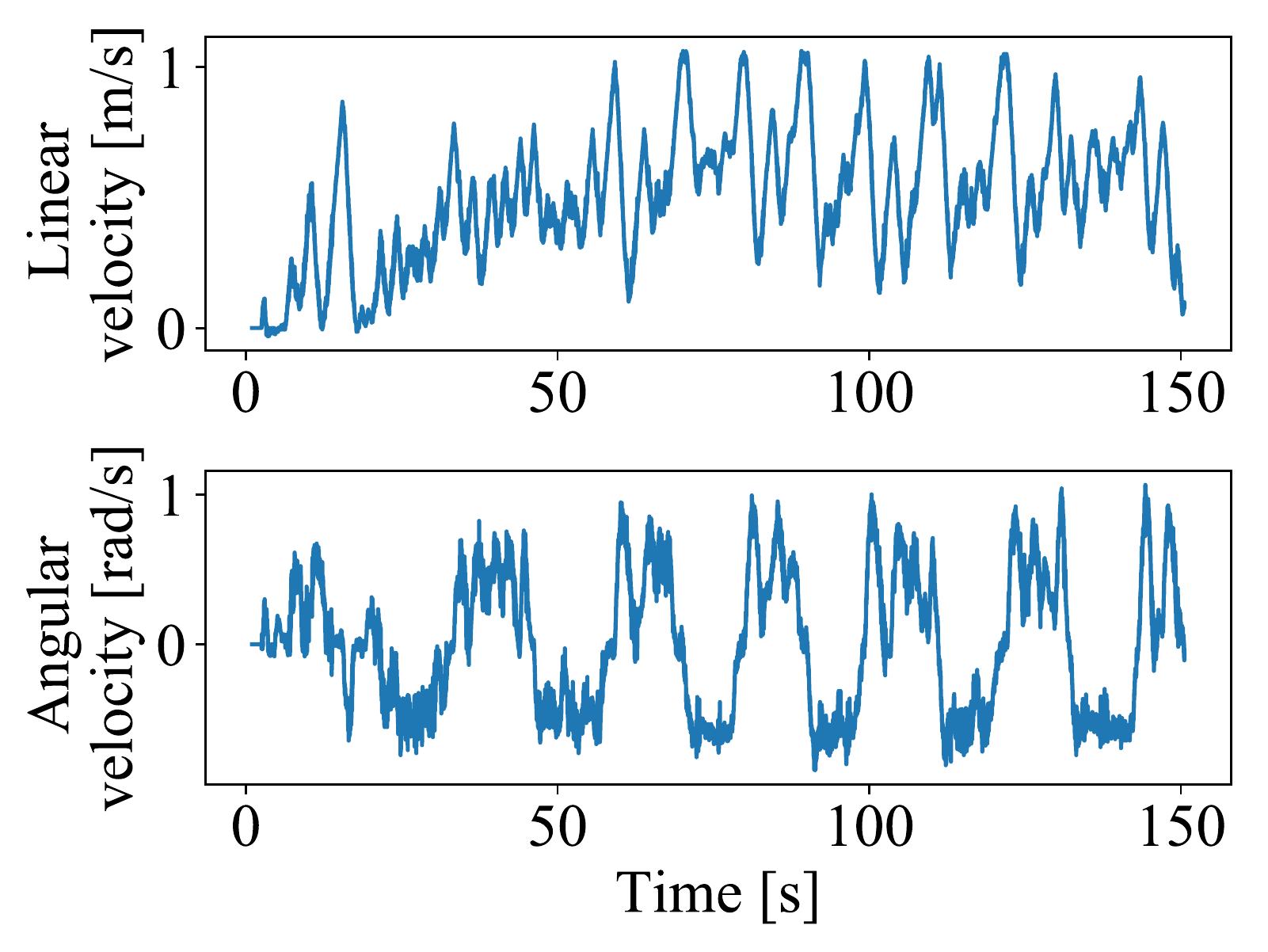}%
		\label{fig:vel_t}}
      \caption{An example of recorded linear/angular velocity (P5, Session2) during the comparative experiment.
      \label{fig:vel} }
   \end{figure}

\begin{figure}[!t]
      \centering
		\subfigure[Session 1]{\includegraphics[width=6.5cm]{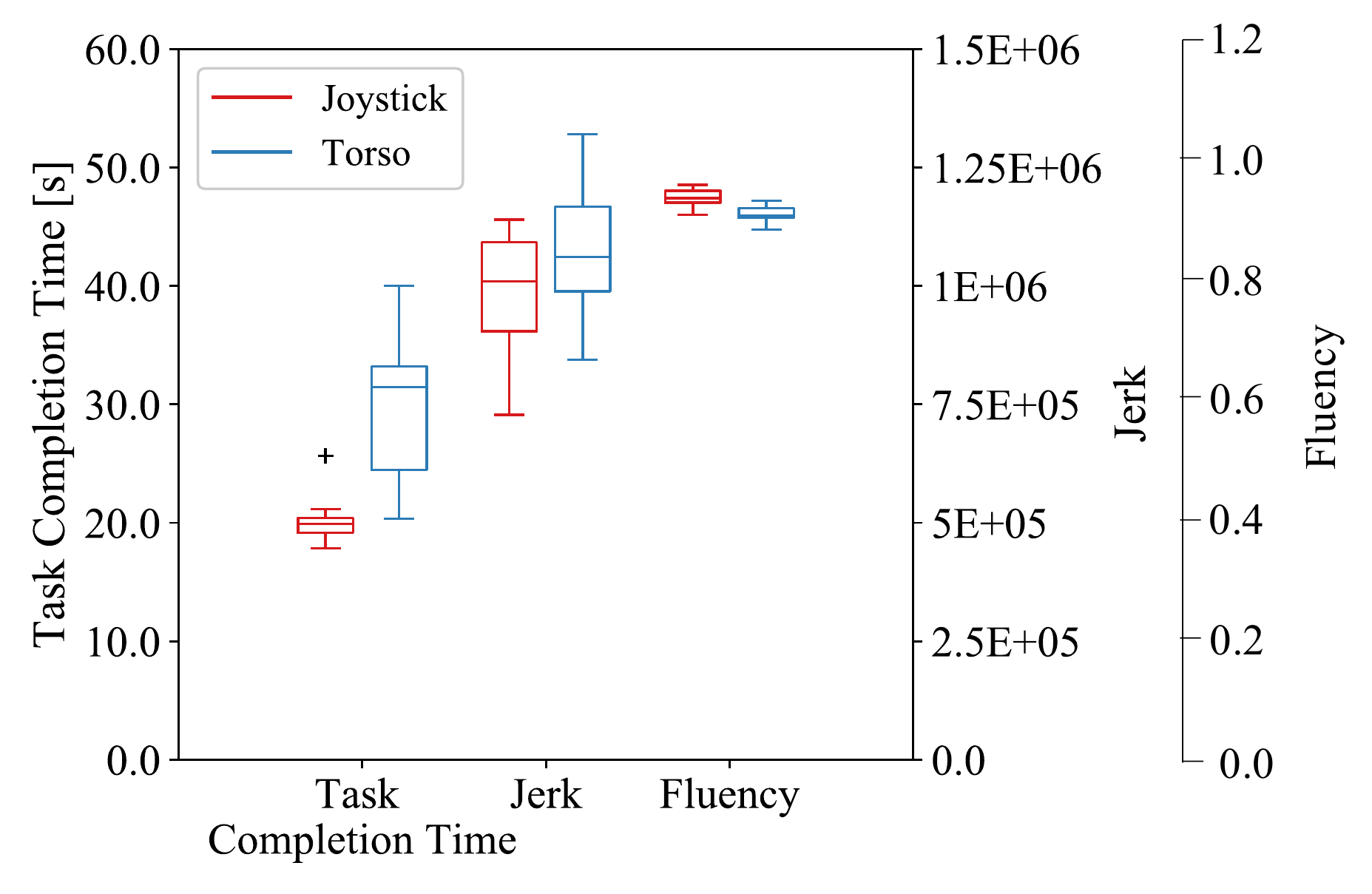}%
		\label{fig:session1}}
		\hfil 
		\subfigure[Session 2]{\includegraphics[width=6.5cm]{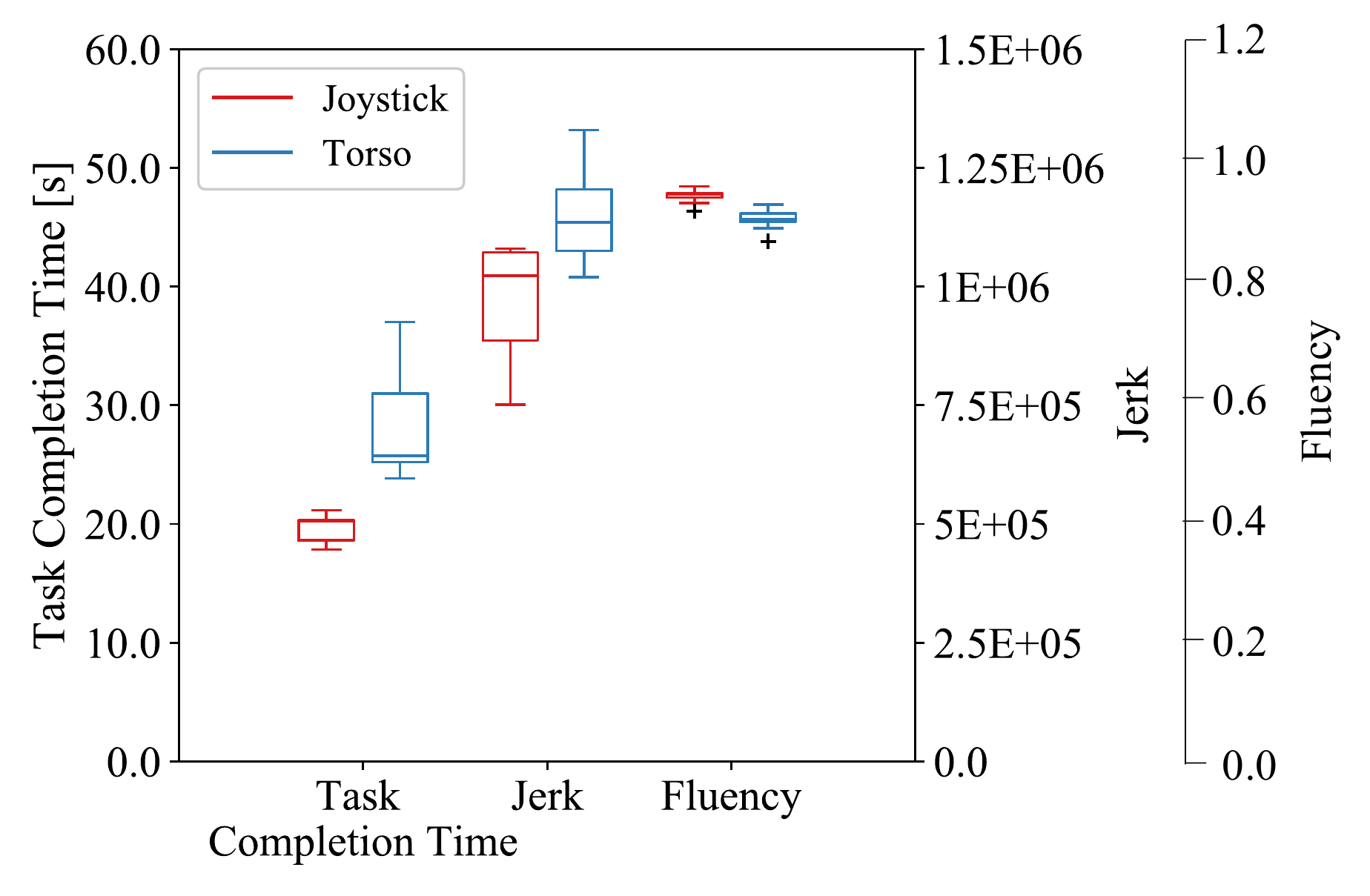}%
		\label{fig:session2}}
      \caption{Statistical analysis of quantitative comparison the proposed torso HMI with a standard joystick for powered wheelchairs.
      \label{fig:3measure} }
   \end{figure}

Recorded velocities of Qolo were used for quantitative evaluation, as shown in Fig. \ref{fig:vel}, with an example from P5, Session 2. Results of statistical analysis of a Wilcoxon signed rank test comparing tasks shows that there was a significant difference (at the level $p<0.05$) between joystick control and embodied torso control in the first session for the task completion time ($p<0.0078$) and fluency of commands ($p<0.0156$), as shown in Fig. \ref{fig:3measure}. With lower times for the joystick interface $20.33\pm 2.3$ against $29.96\pm 6.6$, and higher fluency $94.8 \pm 1.7$ against $92.9 \pm 1.5$. Meanwhile, the jerk over the tests did not show significant difference between the 2 control methods ($p<0.2500$).
On the other hand, session 2 showed a significant difference at the level $p<0.05$, for all three metrics (CT at $p<0.0078$, Fl at $p<0.0469$, and Jk at $p<0.0078$), lower times for the joystick interface $19.6\pm 1.1$ against $28.0\pm 4.5$ with differences in Jerk $20.2\%$ up, and fluency $4.4\%$ down.

Generally, the torso interface takes a longer time than the joystick for completing the task. However, an improvement in the CT observed for most participants suggests to us that there is a learning effect in the torso control. Moreover, for the first session, there was no difference in the overall trajectory jerk, which points out that the users are equally capable of controlling for the task in the narrow space with either interface.

\subsection{User Perception Feedback Analysis}
We designed a questionnaire with aim to measure the subjective perceptions of participants. After the experiment trails, a questionnaire was distributed to each participant to investigate their opinions of torso and joystick control separately. Each questionnaire contains two parts:
\begin{itemize}
 \item The first part asks the participant to give degrees to related perception sentences such as 'I would like to use the robot frequently', which is used to evaluate the system usability score (SUS) \cite{brooke1996quick}.
 \item The second part asks the participant to rate his/her impression of the robot in scales of anthropomorphism, animacy, likeability, perceived intelligence and safety, the so-called "GodSpeed" questionnaire \cite{Bartneck2009}.
\end{itemize}

The questionnaire results are shown in Fig. \ref{fig:quest}: 
\begin{figure}[!t]
    \centering
        \includegraphics[width=6.5cm]{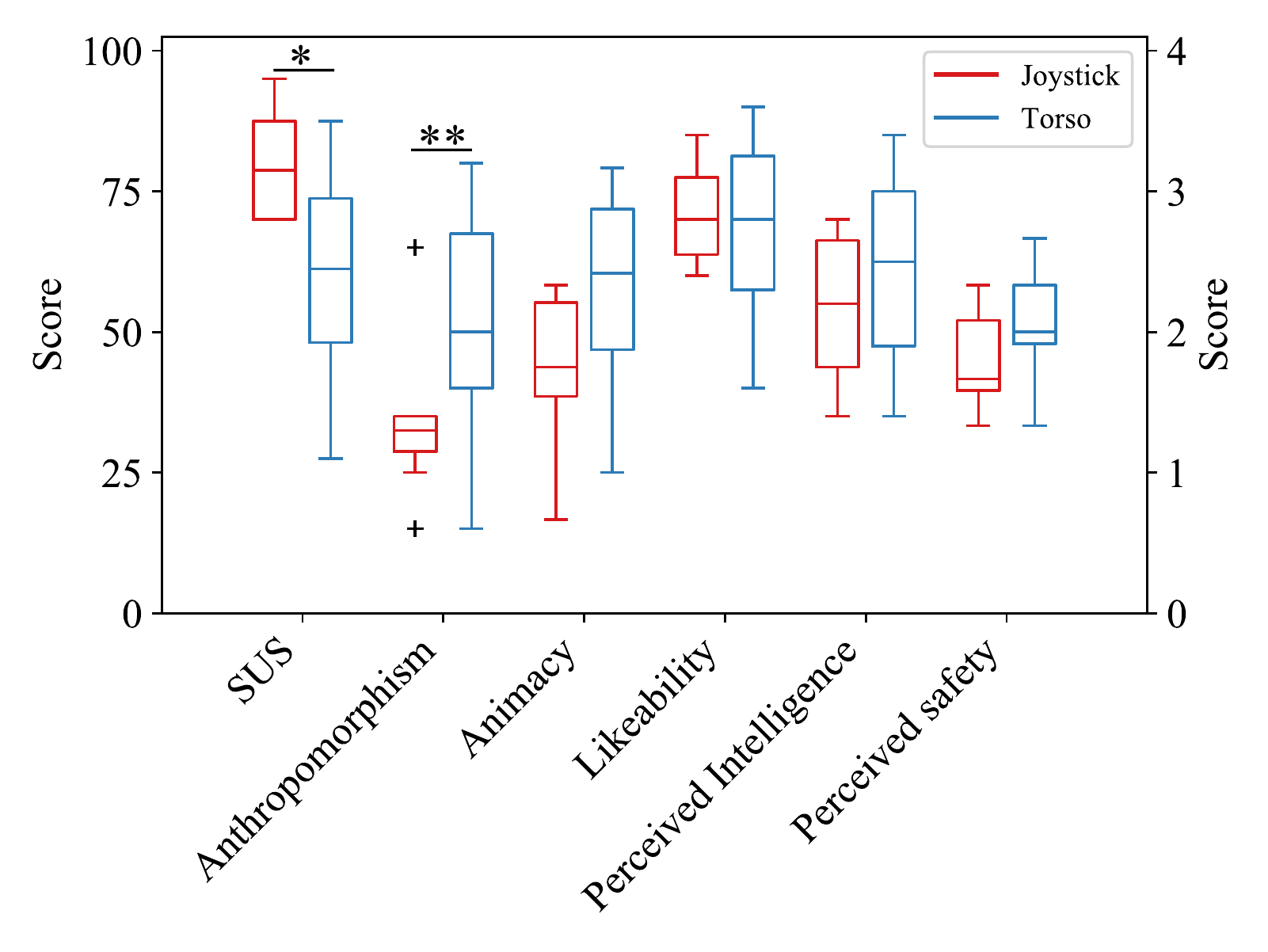}
        \caption{Participants' perception scores of torso/joystick control, with $\star$ denotes significant difference at the level $p < 0.05$, and $\star \star$ at $p < 0.1$.}
\label{fig:quest}
\end{figure}

A Wilcoxon signed rank test was applied for assessing the users' perception difference (with significance level at $p<0.05$) between torso and joystick interfaces.
The result (\(p<0.0312\)) from the perception of System Usability Score (SUS) indicates that the test fails to reject the null hypothesis of zero median in the difference between torso and joystick control at the $5\%$ significance level. Meaning that there was a significant difference in the users' perception with a higher score for a joystick (mean 80), thus,  perceived as more usable against the embodied torso control (mean 62).
On the other hand, perception of anthropomorphism showed significant difference at the $p<0.1$ level (\(p<0.0938\)), being consistently perceived as more natural to us the torso control interface. 
Other perceptions: animacy, likeability, perceived intelligence, and perceived safety resulted in (\(p<0.2188, p<0.6172, p<0.7266, p<0.2500\)), indicating that the null hypothesis of zero median is rejected, therefore, there was not a significant difference between torso and joystick control.
However, as depicted in Fig. \ref{fig:quest} most users found the torso control more animated, intelligent, and even safer.

\section{Conclusions and Future Work} \label{sec_con}
In this work, we proposed and evaluated a novel embodied interface with original control and calibration algorithm, that allows operating a standing mobility vehicle - Qolo - with hands-free control. As such, the proposed interface is proven to be usable by users with control of their torso muscle, specifically, Rectus Abdominis at thoracic region, and Erector Spinae at lumbar region. However, other muscle groups in the torso area could as well replace the function of controlling depending on the user's self selected motor control given their remaining upper-body ability.
We conclude that some degree of control of their abdominal and lumbar muscles could provide enough motion for the current control algorithm and HMI to be used. 

Such insight into the usability of the interface aims to its application for different lower-body impaired users who retain upper-body control (chest, abdominal, and upper limbs), such as, complete and incomplete SCI (we expect users with paraplegia at levels T10 or below), Cerebral Palsy, post stroke wheelchair users and similar lower limbs paralysis.

As expected the joystick performance for the motion task outperformed the torso interface especially in completion time (CT) for all users (as well as in usability perception). However, this equally highlights the potential of the proposed interface, as for all novice users the CT difference was less than $40 \%$, which is a value that could be overcome with practice, as all users were accustomed to the joystick which is a very known HMI.
Moreover, the feedback from participants showed high acceptance in terms of how anthropomorphic the interface could be, as well as animated and even safe for standing locomotion considering that a sense of safety is given by the hands-free control.

Currently, evaluation and preliminary assessment with end-users is being performed and future work will focus on exhaustive assessment of potential end-users and the effectiveness of the proposed interface and algorithms for each of the considered lower-body impairments. What's more, further work should address a wider range of BMI and female users, with perimeter and stiffness of waistline would differ from the male participants in this study.

\addtolength{\textheight}{-12cm}  


%

\section*{ACKNOWLEDGMENT}
This work was partially supported by Grant-in-Aid for Scientific Research from the Ministry of Education, Culture, Sports, Science and Technology of (MEXT) Japan, and the Toyota Mobility Foundation (TMF) by the Unlimited Mobility Challenge grant.


\bibliographystyle{IEEEtran}
\bibliography{IEEEabrv,qolo_mobility_2020}
\end{document}